\documentclass[]{fairmeta}
\usepackage{makecell}
\usepackage{wrapfig}
\usepackage{tabularx}
\usepackage{textcomp}
\usepackage{stfloats}
\usepackage{url}
\usepackage{verbatim}
\usepackage{titlesec}
\usepackage{tocloft}
\usepackage{adjustbox}
\usepackage{multirow}
\usepackage{pifont}
\usepackage[sc]{mathpazo}
\usepackage{tikz}
\usepackage{comment}
\usepackage{amsmath,amssymb}
\usepackage{colortbl}
\usepackage{natbib}
\usepackage{color}
\usepackage{booktabs} 
\usepackage{hyperref}
\usepackage{graphicx}
\usepackage{subcaption}
\RequirePackage{xspace}
\makeatletter
\DeclareRobustCommand\onedot{\futurelet\@let@token\@onedot}
\def\@onedot{\ifx\@let@token.\else.\null\fi\xspace}
\usepackage[most]{tcolorbox}
\usepackage{array}
\usepackage{siunitx}
\usepackage[table]{xcolor}
\usepackage{caption}
\definecolor{headerpurple}{HTML}{d8d2fc}
\definecolor{rowgray}{gray}{0.95}
\usepackage{CJKutf8}

\makeatother

\definecolor{adptorange}{RGB}{248, 205, 172}
\definecolor{cmpblue}{RGB}{189, 215, 238}

\definecolor{our_red}{RGB}{232,157,160}
\definecolor{our_blue}{RGB}{136,206,230}
\definecolor{our_orange}{RGB}{246,200,168}
\definecolor{our_green}{RGB}{178,211,164}

\definecolor{attn_code0}{RGB}{247,215,200}
\definecolor{attn_code1}{RGB}{238,169,139}
\definecolor{mlp_code0}{RGB}{204,201,221}
\definecolor{mlp_code1}{RGB}{102,95,153}
\definecolor{mygray}{HTML}{f0f0f0}

\definecolor{token_blue}{RGB}{84, 120, 140}

\usepackage{bbding}
\usepackage{fontawesome}
\usepackage{float}

\newlength\savewidth

\newcolumntype{x}[1]{>{\centering\arraybackslash}p{#1pt}}
\newcolumntype{y}[1]{>{\raggedright\arraybackslash}p{#1pt}}
\newcolumntype{z}[1]{>{\raggedleft\arraybackslash}p{#1pt}}

\renewcommand{\paragraph}[1]{\vspace{1.25mm}\noindent\textbf{#1}}

\usepackage{algorithm}
\usepackage{listings}

\definecolor{codeblue}{rgb}{0.25, 0.5, 0.5}
\definecolor{codekw}{rgb}{0.35, 0.35, 0.75}
\lstdefinestyle{Pytorch}{
    language = Python,
    backgroundcolor = \color{white},
    basicstyle = \fontsize{9pt}{8pt}\selectfont\ttfamily\bfseries,
    columns = fullflexible,
    aboveskip=1pt,
    belowskip=1pt,
    breaklines = true,
    captionpos = b,
    commentstyle = \color{codeblue},
    keywordstyle = \color{codekw},
}

\definecolor{green}{HTML}{009000}
\definecolor{red}{HTML}{ea4335}

\title{AlayaWorld: Interactive Long-Horizon World Modeling - Full Technical Report}
\author{AlayaWorld Team, Alaya Lab *}
\abstract{
Unlike conventional video game development, which relies on labor-intensive pipelines for asset production, animation, physics, and programming, video world models generate interactive environments from user inputs instantly. It enable us to create customized, explorable, and continuously evolving virtual world from text, an image, or video.
Realizing this vision requires four tightly coupled capabilities: interaction, persistent spatiotemporal consistency, stable long-horizon generation, and efficient response.
We present \textbf{AlayaWorld}, an interactive long-horizon video world model that generates 24-fps video at 540p and 720p.
Built on a 15B video diffusion transformer, AlayaWorld generates short latent chunks autoregressively under camera trajectories and switchable text prompts.
Its bounded visual context combines a persistent sink frame, compressed temporal history, geometry-aligned spatial memory, and recent-frame conditioning.
To reduce long-term drift, the model is trained with corrupted histories and prediction residuals collected from its own roll-outs.
We further introduce a discrete autoregressive distillation formulation that combines distribution-matching distillation, self-forcing++, and consistency distillation, reducing inference from approximately 30 sampling steps to four steps per chunk.
On iWorld-Bench, AlayaWorld achieves the best performance over long-horizon generation.
Conceived as a full-stack, open-source, and long-term project, AlayaWorld is intended to provide an extensible foundation for future research on interactive video world models.
}

\github{\url{https://alaya-lab.github.io/AlayaWorld/}}
\Code{\url{https://github.com/AlayaLab/AlayaWorld}}
\video{\url{https://www.youtube.com/watch?v=n0jIEg7taTI}}

\contact{kaipeng.zhang@shanda.com}
\date{\today}

\begin{document}
\maketitle

\begingroup
\renewcommand{\thefootnote}{}
\footnotetext{* Alaya Lab contributors are listed at the end of the report.}
\endgroup

\begin{figure}[!h]
    \centering
    \includegraphics[
        width=\linewidth,
        trim={0 30mm 0 30mm},
        clip
    ]{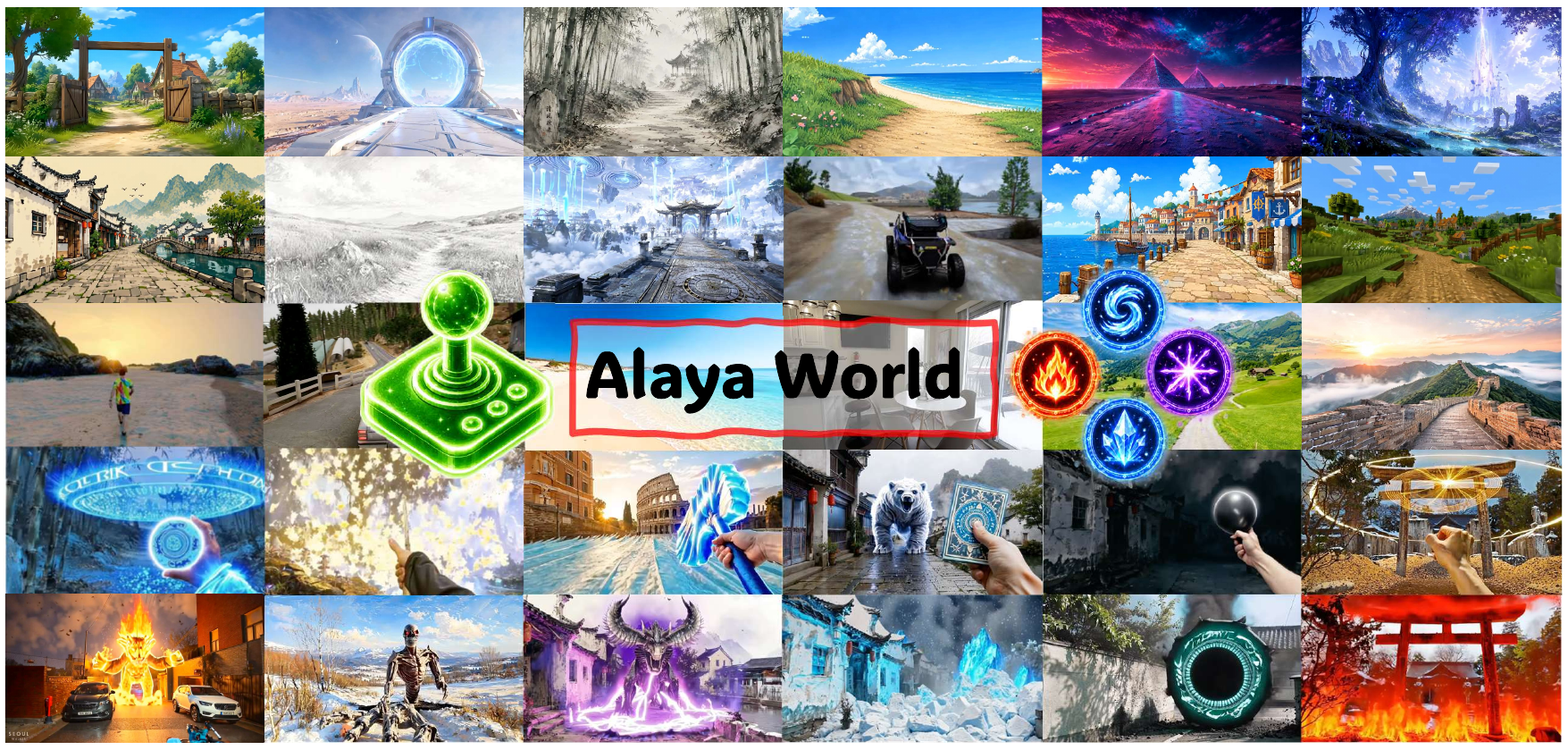}
    \caption{Interactive world simulation across diverse scenes.}
    \label{fig:stage}
\end{figure}

\section{Introduction}
\label{sec:introduction}

\noindent
\textit{\textbf{Reality be rent. Synapse break. Banishment, this world!}}
\par

\noindent\hfill
\textit{--- Watch Love, Chunibyo \& Other Delusions}
\begin{CJK}{UTF8}{min}
\end{CJK}

The dream of engaging with a reality unconstrained by physical laws has captivated humanity for generations. Among existing forms of entertainment, video games have come closest to realizing this vision by creating interactive virtual worlds that respond dynamically to player actions.
However, modern 3D game development depends on a prolonged and tightly coupled production pipeline involving concept design, asset creation, animation, rendering, physics, gameplay programming, testing, and optimization.
The resulting time and labor costs make personalized, rapidly evolving, and effectively unbounded interactive worlds prohibitively expensive.

Video world models offer a fundamentally different route
\cite{mao2025yume,Mao_2026_CVPR,he2025matrix,hyworld2025}.
Rather than manually creating assets and programming interaction rules, such models generate virtual worlds directly from text, images, videos, and user controls.
Visual appearance, motion, viewpoint changes, and certain forms of interaction are encoded jointly in the model parameters and realized through a continuous generative process.

Turning a video generator into a genuinely interactive world model, however, requires four tightly coupled capabilities.
\emph{Interaction} requires the model to respond accurately to camera trajectories and evolving user intent.
\emph{Consistency} requires the world to preserve its spatial structure and visual identity across viewpoint changes and delayed revisits.
\emph{Stability} requires long autoregressive roll-outs without progressively accumulating blur, illumination shifts, or geometric drift.
\emph{Efficiency} requires sufficiently low generation latency and rapid response to new control signals.
These capabilities cannot be addressed independently: broader interaction makes consistency harder to preserve, longer roll-outs amplify residual errors, and aggressive acceleration may further compromise visual stability.

In this work, we present \textbf{AlayaWorld}, an interactive long-horizon video world model.
AlayaWorld is built on a 15B video diffusion transformer and generates 24-fps video at 540p and 720p.
It synthesizes the world autoregressively in short latent chunks under continuous camera trajectories and dynamically switchable text prompts, supporting both controllable navigation and prompt-driven open-ended actions.
To preserve scene information over long roll-outs, AlayaWorld uses a bounded visual context that combines a global scene anchor, compressed recent history, geometry-aligned spatial memory, and recent-frame conditioning.
Temporal memory maintains local dynamics and frame-to-frame continuity, while spatial memory stores past observations and reprojects them into the target view when previously visited regions are revisited.
Because the context size remains bounded, the model generates each new chunk with approximately constant computational cost as the roll-out grows.

AlayaWorld is trained in three stages.
We first adapt a general bidirectional video prior to world-modeling data composed of real-world videos, gameplay recordings, and generated events.
The model is then converted into an autoregressive generator equipped with camera control and spatiotemporal memory.
To mitigate long-horizon drift, we train with corrupted histories and replay prediction residuals collected from the model's own roll-outs, teaching it to recover from imperfect context.
Finally, we introduce a discrete autoregressive distillation method that combines distribution-matching distillation, self-forcing++, and consistency distillation, reducing inference from approximately 30 sampling steps to four steps per chunk while preserving the complete control and memory stack.

We evaluate AlayaWorld on iWorld-Bench
\cite{fang2026iworld}
across generation quality, trajectory following, and memory ability.
AlayaWorld achieves the best overall performance.
Qualitative results further demonstrate controllable navigation, consistent revisitation, prompt-driven actions, and stable long-horizon generation across diverse scenes.

AlayaWorld still represents the world primarily through visual observations, estimated geometry, and visual memory.
Its understanding of object state, physical causality, and long-term task structure therefore remains limited to their visible consequences.

Conceived as a full-stack, open-source, and long-term project, AlayaWorld is intended to provide an extensible foundation for future research on interactive video world models.
\section{Training Data}
\label{sec:data}

Data is the interface through which a video world model learns two coupled
capabilities: photorealistic scene evolution and controllable camera-conditioned
navigation. We therefore construct a corpus that is deliberately heterogeneous
along three axes: visual domain, motion geometry, and supervision fidelity. The
corpus combines \emph{real-world captures}, which anchor the model in natural
appearance, scene layout, and capture artifacts, with \emph{synthetic
renderings}, which provide scalable access to controlled camera motion,
long-tail interactions, and action-driven dynamics. This design follows recent
video world-model pipelines that mix real, synthetic, and action-conditioned
data to improve both appearance diversity and
controllability~\citep{li2025sekai,mao2025yume,he2025cameractrl2}.

\begin{figure*}[t]
\centering
\includegraphics[width=\textwidth]{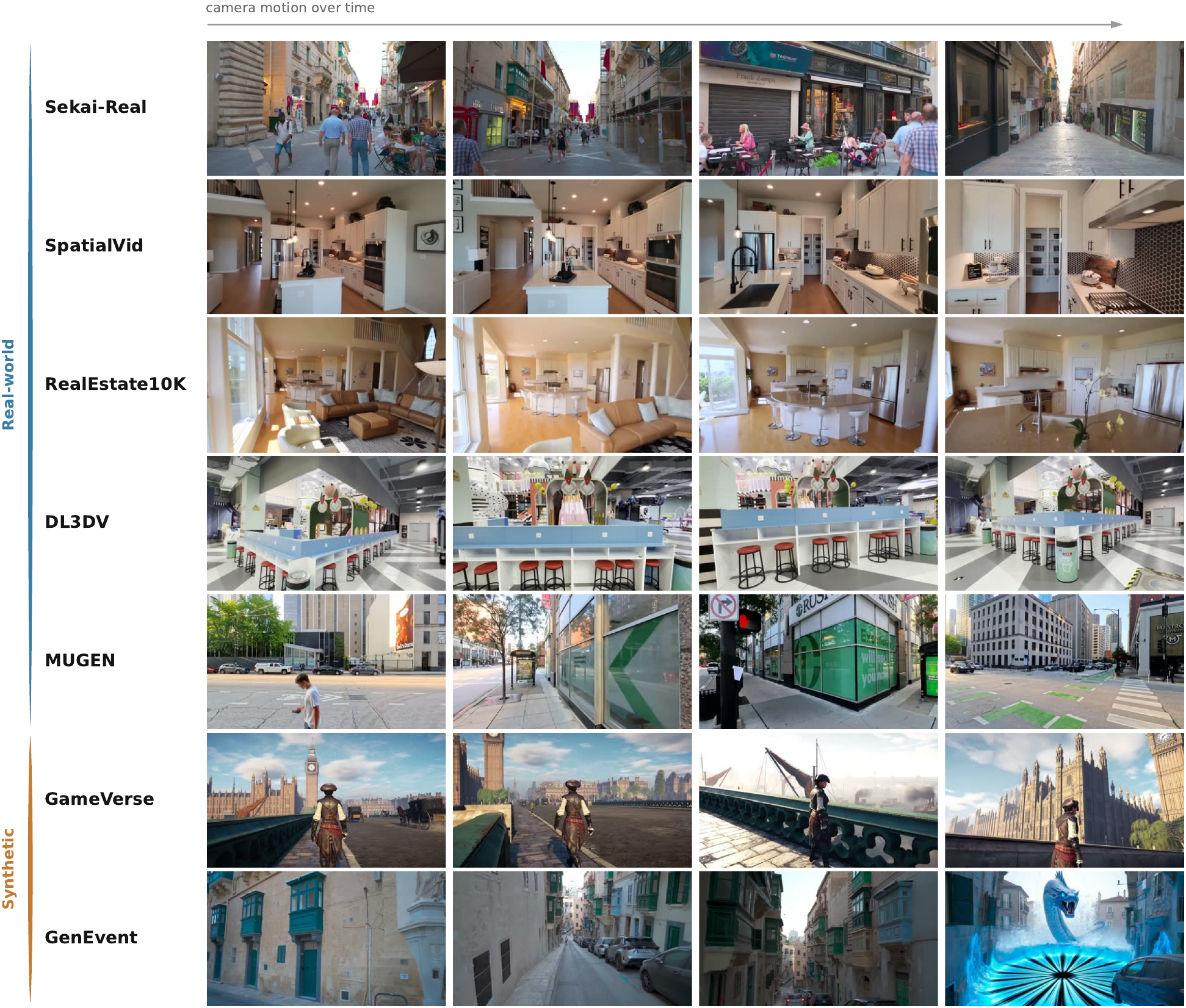}
\caption{Representative training samples from the seven data sources, grouped
by source type. Each row contains four temporally ordered frames from a single
clip, making both the appearance diversity of the corpus and the camera motion
available for supervision visible within one view.}
\label{fig:data-samples}
\end{figure*}

All sources are normalized into a single training record consisting of a video,
per-frame camera intrinsics and pose when available or recoverable, and a
hierarchical caption aligned to the clip timeline. The resulting corpus contains
\num{222147} clips from seven sources, including two internally curated sources
(MUGEN and GameVerse), summarized in
Table~\ref{tab:data-mixture}. Figure~\ref{fig:data-samples} illustrates the
same mixture qualitatively. Each row shows temporally ordered frames from one
clip, so the row should be read as a short trajectory rather than as independent
images. Horizontally, the samples expose the camera motions the model is trained
to follow: forward walking, indoor traversal, panoramic sweeps, third-person
navigation, and event-centric viewpoint changes. Vertically, they show the
domain coverage induced by the mixture, spanning crowded streets, domestic
interiors, real-estate walkthroughs, panoramic reprojections, game-engine
renderings, and generative event videos.

\subsection{Data Mixture}
\label{sec:data-composition}

\paragraph{Real-world captures.}
The real-world portion of the corpus supplies physical appearance, natural
camera shake, sensor artifacts, and scene layouts that are difficult to cover
with simulation alone. We aggregate five complementary sources.
\textbf{Sekai-Real}~\citep{li2025sekai} contributes first-person urban walking
trajectories; \textbf{SpatialVid}~\citep{wang2025spatialvid} adds short clips
with dense indoor camera motion; \textbf{RealEstate10K}~\citep{zhou2018realestate10k}
provides real-estate walkthroughs with broad indoor coverage; and
\textbf{DL3DV}~\citep{ling2024dl3dv} contributes long, contiguous multi-view
walkthroughs of real environments. We keep DL3DV as a standalone source because
its clips are scene-level traversals rather than fixed-length Internet-video
segments. 
Finally, \textbf{MUGEN} is our internally curated dataset, built from YouTube videos, with various camera trajectories and accurate annotation.
For all real-world sources without camera metadata, we recover
per-frame intrinsics and poses using ViPE~\citep{huang2025vipe};

\paragraph{Synthetic renderings.}
The synthetic portion complements real captures with camera and action control
that is difficult to obtain at Internet scale. \textbf{GameVerse} is our
internal large-scale gameplay corpus, which contains \num{124116} clips of approximately \SI{66}{\second} each.
\textbf{GenEvent} contributes
\num{6490} event-centric clips, averaging \SI{10.6}{\second}, synthesized by a
generative video model. These clips expose the model to open-domain,
action-triggered events that are underrepresented in both curated real-world
captures and game footage.

\begin{table}[t]
\centering
\small
\caption{Training data composition. The mixture balances real captures for
photorealism and geometry with synthetic renderings for scalable camera and
action controllability. $^\dagger$ denotes internally curated sources.}
\label{tab:data-mixture}
\begin{tabular}{llr}
\toprule
Source & Type & \#Clips \\
\midrule
Sekai-Real (walking) & real, FPV                 & \num{21561}  \\
SpatialVid           & real, indoor              & \num{23210}  \\
RealEstate10K        & real, indoor              & \num{17429}  \\
DL3DV                & real, walkthrough         & \num{7905}   \\
MUGEN$^\dagger$      & real, FPV & \num{21436}  \\
\midrule
GameVerse$^\dagger$  & synthetic, game    & \num{124116} \\
GenEvent$^\dagger$             & synthetic, event     & \num{6490}   \\
\midrule
Total                &                            & \num{222147} \\
\bottomrule
\end{tabular}
\end{table}

\subsection{Curation Pipeline}
\label{sec:curation}

A heterogeneous corpus is only useful if quality decisions are made
consistently across sources. We therefore use a unified, stage-based filtering
pipeline rather than source-specific heuristics. The pipeline operates over a
shared manifest in three phases: \texttt{ingest} builds the manifest,
\texttt{run} evaluates all filtering stages over a dependency DAG, and
\texttt{select} applies the final all-stages-must-pass rule. Expensive features
are computed once, cached in the manifest, and reused by downstream gates;
optional global \texttt{rank-cut} and near-duplicate \texttt{dedup} stages can
be inserted before final selection.

\paragraph{Shared feature cache.}
The filtering backbone uses a single-decode design. For each clip, one NVDEC pass decodes frames and one RAFT~\citep{teed2020raft} forward pass estimates optical flow. From these outputs we compute frame-level summaries, including luminance, frame differences, and border statistics, together with flow-derived motion features such as flow-per-second, temporal variance, and directional variance. All rule-based stages read these cached scores at millisecond cost, avoiding repeated video decoding and making large-scale threshold sweeps practical.

\paragraph{Filtering stages.}
Each clip must pass six classes of gates. First, \emph{technical validation}
checks decode integrity, minimum resolution ($\geq\!720$p), duration
($\geq\!3$\,s), frame rate ($24$--$65$\,fps), and codec membership
(H.264/HEVC/AV1). Second, \emph{photometric validation} rejects clips with
extreme exposure or excessive black borders, using a maximum border ratio of
$0.10$. Third, \emph{shot-boundary filtering} combines classical
cut/dissolve detection with OmniShotCut~\citep{wang2026omnishotcut}, ensuring
that each retained training sample is a single continuous shot. Fourth,
\emph{motion analysis} removes static clips, scores temporal consistency,
assigns camera-motion buckets (static, pan, gameplay, mixed), and detects
pose-free camera shake. Fifth, \emph{text and interface suppression} removes
subtitles, watermarks, and game HUD overlays using EasyOCR-based text
detection~\citep{jaidedai2020easyocr} and a pixel-stability UI mask with
maximum overlay ratio $0.04$. Sixth, \emph{person control} applies a
YOLO11~\citep{khanam2024yolo11} detector to bound foreground-human count and
screen occupancy.

For sources with estimated camera trajectories, we additionally apply a
\emph{pose stability} gate. The gate scores trajectory jitter, peak
acceleration, median reconstruction residual, and long-horizon drift, with
source-specific anchors because walking videos and game footage occupy
different motion scales. Perceptual-quality estimators
(COVER~\citep{he2024cover}, VBench~\citep{huang2024vbench}) and
SigLIP2/CLIP/V-JEPA2 embeddings~\citep{tschannen2025siglip2,radford2021clip,assran2025vjepa2}
support the optional global \texttt{rank-cut} and near-duplicate
\texttt{dedup} stages.

\paragraph{Selection protocol.}
The final training set consists only of clips that pass every active gate; these
surviving clips form the \texttt{*\_filtered} splits reported in
Table~\ref{tab:data-mixture}. We calibrate thresholds per source by profiling a
$\sim\!200$-clip slice before enabling a gate, since score distributions differ
substantially across capture conditions and game genres. This calibration makes
filtering strict enough to remove failure modes while preserving source-specific
motion statistics.

\subsection{Training-Oriented Caption Annotation}
\label{sec:caption}

A single global caption is insufficient for training a controllable world model:
at inference time the model receives localized, time-varying instructions, while
a clip-level caption provides no aligned temporal supervision. We therefore
annotate each clip with a two-level schema that explicitly separates global
context from segment-level dynamics. Captions are generated by a vision-language
model, with Kimi-K2.6~\citep{kimiteam2026kimi26} as the default backend and
Gemini~\citep{comanici2025gemini25} and
Gemma~\citep{gemmateam2025gemma3} as alternatives. The model consumes frames
sampled at $1$--$2$\,fps, each tagged with an explicit \texttt{[mm:ss]}
timestamp, which encourages temporal segmentation rather than a single
collapsed description.

\paragraph{Video-level context.}
At the clip level, we annotate a compact set of global attributes: weather,
time of day, location type, camera perspective, camera motion, and video style.
These attributes are drawn from a deliberately small vocabulary, reduced from
59 to 26 values, so they can serve both as conditioning tokens and as reliable
keys for data balancing, for example by \texttt{location\_type} or
\texttt{camera\_perspective}.

\paragraph{Segment-level tracks.}
Each clip is partitioned into timestamped segments, represented by
\texttt{time\_range\_s}. Within each segment, we annotate separate semantic
tracks rather than a single entangled sentence: primary subject motion,
environmental dynamics, static scene attributes, and camera viewpoint/motion
(Table~\ref{tab:caption-tracks}). The separation between subject and camera
tracks is crucial. It allows the annotation to distinguish, for example, a
walking subject from a dolly-in camera, an orbit from a turning agent, or a
static subject under a panning viewpoint. The descriptive tracks are fused into
a natural-language \texttt{full\_prompt} of 5--9 present-tense sentences for the
text encoder, together with a 15--45-word \texttt{short\_prompt} used for
caption dropout and multi-caption augmentation.


\begin{table}[t]
\centering
\small
\caption{Per-segment annotation tracks. The descriptive tracks are fused into
\texttt{full\_prompt} and \texttt{short\_prompt}; \texttt{camera\_path}
provides a discrete camera-trajectory control signal independent of subject
motion.}
\label{tab:caption-tracks}
\begin{tabular}{lll}
\toprule
Field & Type & Role \\
\midrule
\texttt{subject\_motion}      & free text & primary-agent motion \\
\texttt{environment\_motion}  & free text & dynamics of entities, lighting, and weather \\
\texttt{static\_scene}        & free text & time-invariant scene attributes \\
\texttt{camera\_description}  & free text & viewpoint, framing, motion, and stability \\
\texttt{full\_prompt}         & free text & fused caption for the text encoder \\
\texttt{short\_prompt}        & free text & compact caption for dropout and augmentation \\
\texttt{camera\_path}         & enum (16) & discrete camera-trajectory control target \\
\bottomrule
\end{tabular}
\end{table}




\section{AlayaWorld}
\label{sec:model}


AlayaWorld is an interactive world model built on top of the LTX-2.3.
The public LTX-2.3 checkpoint is a $22$B multimodal model; we remove its audio module, leaving a $\sim$13B video DiT that constitutes our backbone.

Training proceeds in three stages:
(i)~\emph{bidirectional pre-training}, which adapts the general video prior to our domain with a full-parameter fine-tune;
(ii)~\emph{autoregressive training}, which grafts a history-compression module, a spatial memory, and camera control onto the backbone, and stabilises long roll-outs with anti-drift training;
and (iii)~\emph{post-training acceleration}, which distils the many-step teacher into a $4$-step student via Distribution-Matching Distillation (DMD) combined with self-forcing++ and consistency distillation.

\begin{figure}[t]
    \centering
    \includegraphics[width=0.9\linewidth]{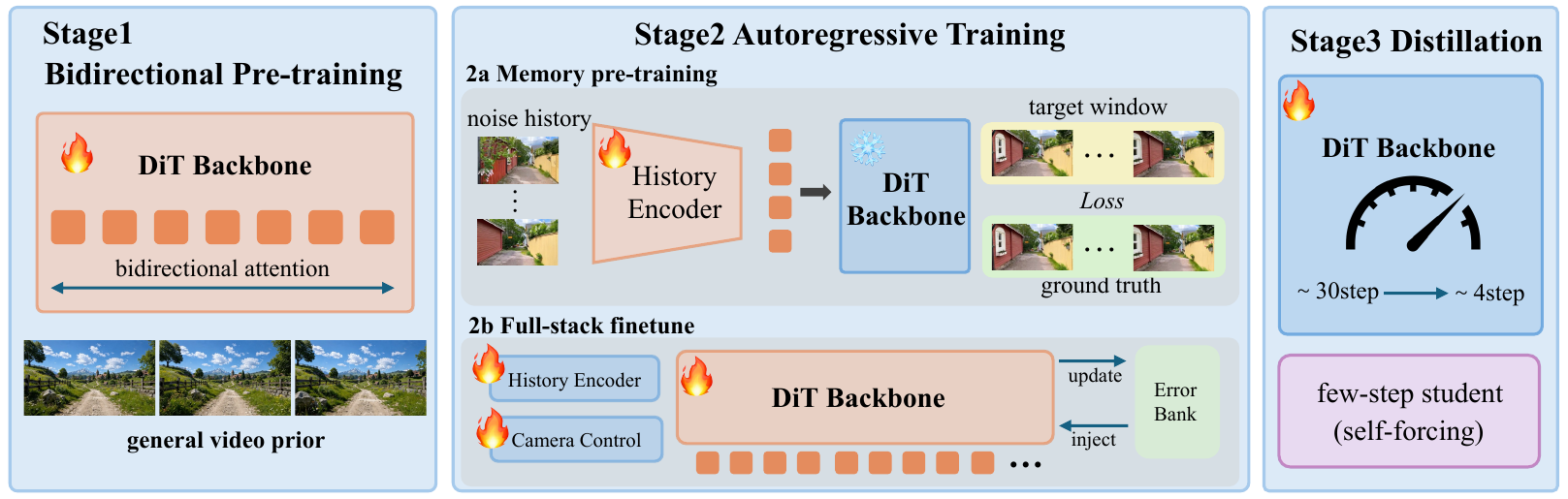}
    \caption{Training stages.}
    \label{fig:stage}
\end{figure}


\subsection{Formulation}
\label{sec:formulation}

\noindent
AlayaWorld generates video \emph{autoregressively in the VAE latent space}, chunk by chunk. The causal video VAE encodes a clip into a latent sequence partitioned into chunks $\{z_1,z_2,\dots\}$, each a block of $K{=}4$ latent frames. At step $i$ the control input is the chunk's target \emph{camera trajectory} $\pi_i$ --- a sequence of absolute camera poses, one per latent frame of the chunk --- together with an optional chunk-level text prompt $y_i$; a prompt switched in at a chunk boundary drives \emph{prompt-driven actions} such as combat or spell-casting. Generation factorises causally as
\begin{equation}
p_\theta\!\left(z_{1:N}\mid \pi_{1:N},y_{1:N}\right)
= \prod_{i=1}^{N} p_\theta\!\left(z_i \mid z_{<i},\, \pi_{\le i},\, y_i\right),
\label{eq:causal}
\end{equation}
so every chunk sees only the past ($z_{<i},\pi_{\le i}$) and never leaks future information. The two conditioning modalities enter by different routes: the camera trajectory is injected as a compact per-frame condition --- the relative pose between consecutive frames --- through an adaptive layer-norm (AdaLN) camera-control module, whereas the visual past $z_{<i}$ enters as an \emph{in-context token prefix}.

\begin{figure}[t]
    \centering
    \includegraphics[width=1\linewidth]{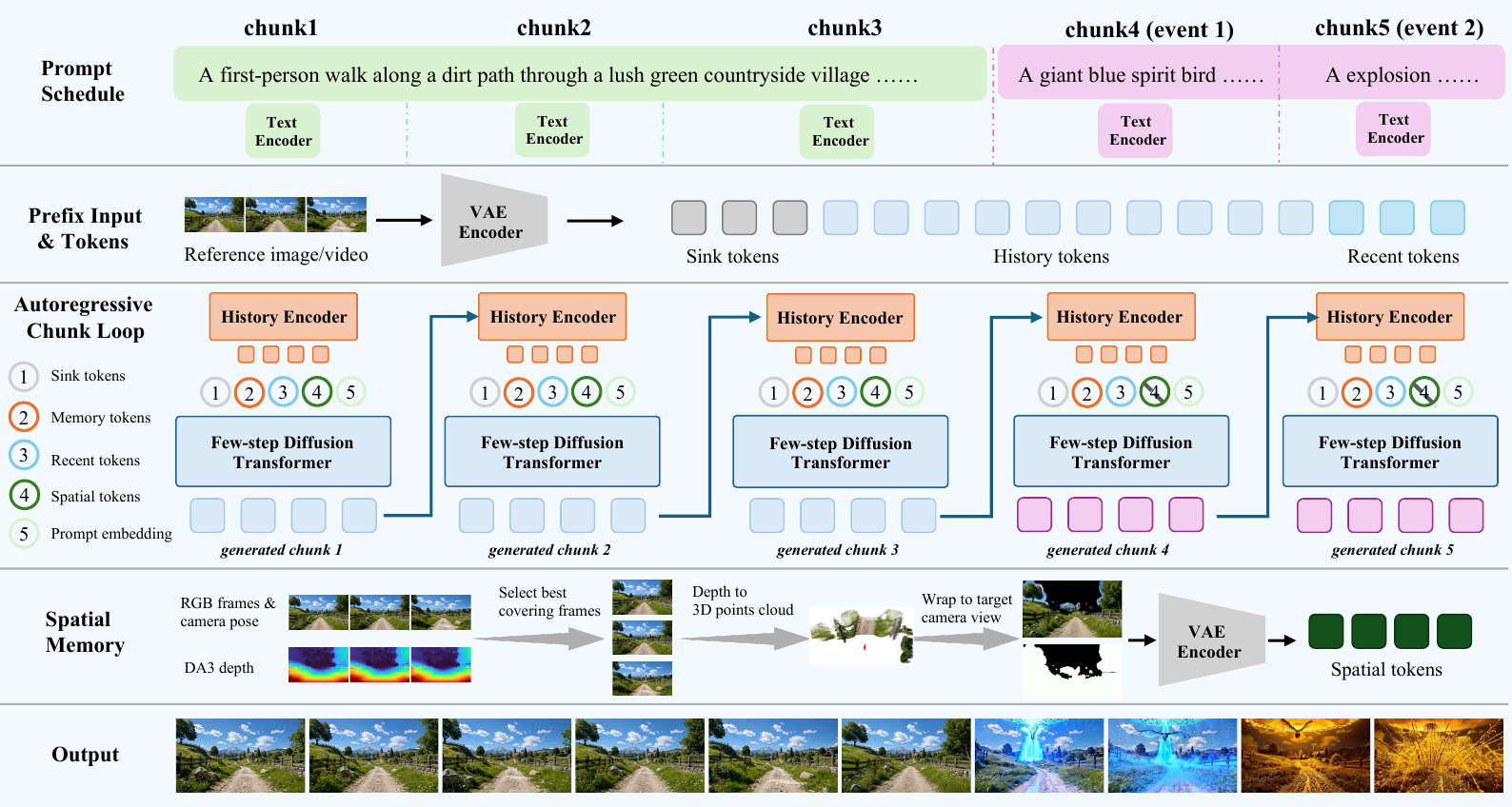}
    \caption{Formulation.}
    \label{fig:formulation}
\end{figure}

\smallskip
\noindent\textbf{Conditioning as an in-context prefix.} The backbone is a single self-attention DiT. The visual context is not supplied through a separate cross-attention bank; instead, for chunk $i$ the model prepends four \emph{clean} ($\sigma{=}0$, noise-free) conditioning streams to the $K$ noised target frames, forming one token sequence
\begin{equation}
S_i = \big[\;
\underbrace{s}_{\text{sink}}\;;\;
\underbrace{h_i}_{\text{temporal memory}}\;;\;
\underbrace{g_i}_{\text{spatial memory}}\;;\;
\underbrace{n_i}_{\text{nearby / I2V}}\;;\;
\underbrace{z_i^{\tau}}_{\text{target}}\;\big],
\label{eq:seq}
\end{equation}
which is processed by full (non-causal) self-attention; the whole prefix is sliced off after the last transformer block and only the target segment is denoised. The four streams differ in source and role:
\begin{itemize}
  \item \textbf{Sink} $s$ --- a \emph{single} clean latent frame, patch-embedded and pinned at RoPE temporal position $0$, held \emph{fixed across all chunks} as a global identity/appearance anchor. During training it is drawn as a \emph{remote} frame (at least $8$ latent frames from the target), which prevents the model from extrapolating the next chunk directly from it and thereby increases its reliance on the camera-control signal.
  \item \textbf{Temporal memory} $h_i=H_\phi(w_i)$ --- a \emph{compressed temporal history}. A history-compression module $H_\phi$ (following Frame Preservation) encodes a \emph{sliding window} $w_i=z_{i-L:i}$ of the last $L{=}6$ latent frames into a lightweight embedding, recomputed every chunk; its tokens are injected directly, bypassing the patch embedder.
  \item \textbf{Spatial memory} $g_i$ --- a geometry-aligned rendering of past views into the current view (Eq.~\eqref{eq:warp}), giving the generator concrete visual evidence for the queried viewpoint.
  \item \textbf{Nearby / I2V condition} $n_i$ --- the \emph{single most-recent} latent frame (the last frame of $w_i$), patch-embedded and placed immediately before the target; this is the image-to-video conditioning frame that carries full-resolution frame-to-frame continuity.
\end{itemize}

\smallskip
\noindent\textbf{Building and rendering the spatial memory.} Following GEN3C, the spatial memory maintains an explicit cache $\mathcal{B}=\{(I_j,D_j,\pi_j)\}$ --- each previously generated frame $I_j$ with its monocular depth $D_j$ (Depth-Anything-3) and camera pose $\pi_j$, keyed by global frame index --- and renders it along the target camera trajectory. The AdaLN camera condition is the per-frame increment of the same trajectory, so the cache and the camera control share one pose source. To render the cache into chunk $i$ (target camera $\pi_i$):
\begin{enumerate}
  \item \textbf{Retrieve} up to $10$ frames by greedy \emph{maximum-coverage} selection: each candidate's depth is unprojected to world points and projected into $\pi_i$; a z-buffer with occlusion tolerance $\delta{=}0.1$ marks which target pixels it covers, and frames are chosen to maximise the number of newly-covered pixels.
  \item \textbf{Warp} each selected frame into the target view by forward splatting,
  \begin{equation}
  u' = \pi_i\!\big(\,\pi_j^{-1}(u,\,D_j(u))\,\big),
  \label{eq:warp}
  \end{equation}
  i.e. unproject pixel $u$ of frame $j$ to a 3D point via its depth and camera, then reproject into $\pi_i$; per-pixel occlusion is resolved by nearest depth and multiple sources are fused into one warped image $\tilde I_i$ plus a binary coverage mask $M_i$.
  \item \textbf{Inject}: $\tilde I_i$ is VAE-encoded to $g_i$ and placed at the target's RoPE coordinates; the coverage mask $M_i$ becomes a self-attention key bias so uncovered (i.e. never-observed) regions are ignored rather than trusted.
  \item \textbf{Update}: once chunk $i$ is generated it is decoded to pixels, its depth is estimated, and $(I_i,D_i,\pi_i)$ is appended to $\mathcal{B}$.
\end{enumerate}
This gives long-range spatial consistency --- revisiting a previously seen place renders it consistently --- well beyond the $6$-frame temporal window. Writing $c_i=(s,h_i,g_i,n_i,\pi_{\le i},y_i)$ for the full per-chunk conditioning --- where the four context streams $s,h_i,g_i,n_i$ are all functions of the past $z_{<i}$ --- the causal factor $p_\theta(z_i\mid z_{<i},\pi_{\le i},y_i)$ of Eq.~\eqref{eq:causal} is realised as $p_\theta(z_i\mid c_i)$.

\smallskip
\noindent\textbf{Roll-out.} One interactive step is therefore: (1)~read the camera trajectory $\pi_i$ (and optional prompt $y_i$); (2)~build the prefix $S_i$ of Eq.~\eqref{eq:seq} --- pin the sink $s$, recompute the temporal memory $h_i$ over the last $6$ latents, render the spatial memory into $g_i$, and set the nearby frame $n_i$; (3)~sample the next chunk $\hat z_i\sim p_\theta(\cdot\mid c_i)$; (4)~stream-decode $\hat z_i$ to pixels, slide the history window forward, and append the new frames to the spatial-memory cache $\mathcal{B}$; (5)~advance $i\leftarrow i+1$. Because the context is a bounded rolling window (a fixed sink, a $6$-frame history, and a capped set of $10$ rendered cache frames), the compute per chunk is constant and the horizon $N$ is in principle unbounded, giving arbitrarily long interactive generation.

\subsection{Bidirectional Model Pre-Training -- Establishing the General Video Prior}

\noindent
The base LTX-2.3 model is a general text-to-video prior. This stage adapts it to our world-model domain through a \emph{full-parameter} fine-tune, so that the whole backbone absorbs the new visual and temporal statistics while retaining the base model's generative prior. The model stays fully bidirectional here; no memory or control mechanism is introduced yet. Fine-tuning runs at $24$~fps under a mix of $540$p and $720$p resolutions, on variable-length clips of up to $20$~seconds --- the temporal range the base model was trained on. To prime the backbone for the conditional generation used later, clips are trained under a mixture of image-, video-, and text-conditioned objectives.

\noindent
The training data is a weighted mixture dominated by a balanced scene / camera-pose corpus, complemented by AAA-gameplay recordings, real first-person walkthroughs, and magic-event clips for prompt-driven actions such as combat, spell casting, and monster summoning; a per-sample weighted sampler balances the dominant source. Text conditioning is chosen by the length of the underlying video segment: segments longer than $20$~seconds are described only by a single high-level whole-clip summary, whereas segments within $20$~seconds randomly alternate between that summary and a detailed per-segment caption. In either case the descriptive content inside the \texttt{<camera>} tag is randomly dropped, so that the model does not over-rely on the textual camera description --- camera control is instead supplied by a dedicated signal in the later stages. We adopt an \emph{adaptive sigma-shift} schedule whose flow-matching timestep shift scales with clip length, allocating the denoising budget appropriately across durations, and close the stage with a short low-$\sigma$ refinement pass that sharpens fine detail.

\subsection{Autoregressive Model Training -- Control and Memory Integration}

\noindent
This stage turns the bidirectional generator into an \emph{autoregressive, controllable} world model that rolls out chunk-by-chunk, conditioned on its own past, a persistent memory, and a user-supplied camera trajectory. It is carried out in two phases with different training regimes.

\noindent
\textbf{History pre-training.} The first phase keeps the backbone \emph{frozen} and trains, via a LoRA adapter, the history-compression module $H_\phi$ --- which, following \emph{Frame Preservation}, compresses the recent frame history into a lightweight embedding. Given a history window $\{z_1,\dots,z_L\}$, we mask it by perturbing each frame with its own noise level, $\tilde z_i=(1-\sigma_i)z_i+\sigma_i\epsilon_i$ with $\sigma_i\sim\mathcal{U}(0.2,1)$, and ask the backbone (a rectified-flow velocity field $v_\theta$) to reconstruct a short target window $z_\Omega$ conditioned on the compressed masked history, under the flow-matching objective
\begin{equation}
\mathcal{L}_{\text{2a}}=\mathbb{E}\big\|\,v_\theta\!\big(z_\Omega^{\tau},\tau \mid H_\phi(\tilde z)\big)-(\epsilon-z_\Omega)\big\|_2^2,
\qquad z_\Omega^{\tau}=(1-\tau)z_\Omega+\tau\epsilon,\ \ \tau\sim\mathcal{U}(0,1).
\label{eq:phase2a}
\end{equation}
The target-window length is varied so the module serves both short and long horizons. This phase uses no control signal, pre-training the memory pathway in isolation.

\noindent
\textbf{Full-stack fine-tuning.} Starting from the history-pretrained weights, the second phase is a \emph{full-parameter} supervised fine-tune: the backbone is unfrozen and trained in full, together with three dedicated modules --- the history-compression module, the camera-control module, and a next forcing head~\cite{xu2026next}. Its components are:
\begin{itemize}
  \item \textbf{Camera control.} Each per-frame relative-pose increment $\Delta\pi$ of the camera trajectory is Fourier-embedded per axis, concatenated over the six pose components, and passed through an MLP; the result modulates the tokens through AdaLN:
  \begin{equation}
  c_{\text{cam}}=\mathrm{MLP}\Big(\textstyle\bigoplus_{k=1}^{6}\mathrm{PE}(\Delta\pi_k)\Big),
  \qquad e \leftarrow e + c_{\text{cam}},
  \label{eq:cam}
  \end{equation}
  where $e$ is the timestep embedding from which the AdaLN scale and shift are produced. Per-axis scales are calibrated from the motion statistics of the real data.
  \item \textbf{Temporal memory.} The history-compression module from the first phase is kept trainable and recomputed each chunk over the sliding history window, with history-dropout for robustness.
  \item \textbf{Spatial memory.} Following GEN3C, an explicit cache stores previously generated frames with their monocular depth (Depth-Anything-3) and camera pose, and renders them into the current view by maximum-coverage retrieval, giving long-range spatial consistency beyond the history window.
  \item \textbf{Next forcing}. An auxiliary head reinforces frame-to-frame causal continuity by predicting the \emph{next} chunk from the backbone's intermediate features: hidden states hooked from several layers are fused into a feature $F$ and, together with the noised next chunk, decoded by a small head $f_\psi$ into a velocity, supervised at a shifted (higher) noise level $\tilde\tau=\tfrac{10\,\tau}{1+9\,\tau}$,
  \begin{equation}
  \mathcal{L}_{\text{nf}}=\big\|\,f_\psi\big(F,\ z^{+,\tilde\tau}_{0},\ \tilde\tau\big)-(\epsilon-z^{+}_{0})\big\|_2^2,
  \qquad \mathcal{L}=\mathcal{L}_{\text{flow}}+0.5\,\mathcal{L}_{\text{nf}},
  \label{eq:nf}
  \end{equation}
  where $z^{+}_{0}$ is the next chunk and $\mathcal{L}_{\text{flow}}$ is the main flow-matching objective of the chunk being generated.
\end{itemize}
The data mixture follows the pre-training stage, with the per-source proportions retuned for this phase.

\noindent
\textbf{Anti-drift training strategy.} Long autoregressive roll-outs accumulate error, so we train the model to tolerate a corrupted past. Both mechanisms below act on the \emph{temporal-memory}, \emph{spatial-memory}, and \emph{nearby} context tokens (the sink is kept clean):
\begin{itemize}
  \item \textbf{Helios drift simulation}~\cite{yuan2026helios} degrades the context in latent space with one of three artefact types that mimic what a roll-out drifts into --- additive noise $z\mapsto(1-\sigma)z+\sigma\epsilon$ (with $\sigma\sim\mathcal{U}(0,\rho)$, $\rho$ a corruption strength), a down/up-sampling blur $z\mapsto\mathrm{up}(\mathrm{down}(z;r))$ ($r\sim\mathcal{U}(0.9,1)$), and a saturation shift $z\mapsto(z-\bar{z})\,\alpha+\bar{z}$ ($\alpha\sim\mathcal{U}(0.3,1.7)$) --- where a noise-or-blur step is optionally followed by a saturation step.
  \item \textbf{Error bank}~\cite{li2025stablevideoinfinity} keeps the model's own reconstruction residuals $\delta=\hat{z}_0-z_0$ (with $\hat{z}_0=z^{\tau}-\tau\,v_\theta$) in a buffer bucketed by chunk length and noise level, and replays them additively into the context (and the target latent), $z\leftarrow z+\gamma\,\delta$, so the model learns to recover from the failure modes it actually produces at inference.
\end{itemize}
The two are scheduled through an error-bank warm-up: before the bank has filled, only Helios drift is applied (at a fixed probability); once the bank is warmed up it takes priority and the Helios probability is lowered, the two being mutually exclusive within any step.

\subsection{Post-Training -- Inference Acceleration}
\label{sec:posttrain}

\noindent
The model from the autoregressive stage is strong but needs many ($\sim$$30$) sampling steps per chunk, too slow for interactive use. We distil it into a \textbf{$4$-step} student. Inspired by the joint consistency-distillation and distribution-matching principle of Causal-rCM~\cite{zheng2026causal}, we introduce a discrete distillation formulation tailored to our autoregressive world model. Unlike the continuous-time formulation of Causal-rCM, our discrete formulation avoids Jacobian-vector-product computation. We further incorporate self-forcing++ to account for the distribution shift induced by autoregressive roll-out. The student retains the full control and memory stack, and is optimized using distribution-matching distillation~\cite{yin2024improved}, self-forcing++~\cite{cui2025self}, and consistency distillation~\cite{song2023consistency}.

\begin{itemize}
  \item \textbf{Distribution-Matching Distillation.} The student is trained to match the teacher's output distribution through a real and a fake (critic) score. Both are served by the \emph{same} score backbone via LoRA swapping rather than a second network --- the critic LoRA \emph{off} gives the real (teacher) score, \emph{on} gives the fake (critic) score --- and the critic is updated more frequently than the student (a two-timescale update rule) so that it stays ahead.
  \item \textbf{Self-forcing++.} Instead of distilling on teacher-forced clips, the student rolls out its own multi-chunk trajectories and is scored against the teacher along that self-generated path (with ground-truth context and detached history). This closes the train/inference gap of autoregressive generation and is the key to seam-free chunk continuation.
  \item \textbf{Consistency distillation.} A consistency loss between adjacent noise levels on a grid of $50$ matches the student's prediction at a higher noise level to an EMA copy of itself at the neighbouring lower level, which stabilises the few-step solution and suppresses brightness/appearance flicker at chunk boundaries.
\end{itemize}
Concretely, DMD pushes the student distribution $p_{\theta,\tau}$ toward the data distribution $p_{\mathrm{data},\tau}$ by the score-difference gradient
\begin{equation}
\nabla_\theta\, D_{\mathrm{KL}}\!\left(p_{\theta,\tau}\,\|\,p_{\mathrm{data},\tau}\right)
= -\,\mathbb{E}\Big[\big(
s_{\mathrm{real}}(\hat z_i^\tau,\tau\mid c_i)
- s_{\mathrm{fake}}(\hat z_i^\tau,\tau\mid c_i)\big)\,
\tfrac{\partial \hat z_i}{\partial\theta}\Big],
\label{eq:dmd}
\end{equation}
where $\hat z_i$ is a chunk from the student's own self-roll-out and the two scores are served by the \emph{same} score backbone with the critic LoRA off ($s_{\mathrm{real}}$, teacher) or on ($s_{\mathrm{fake}}$, trainable critic). The consistency-distillation term enforces trajectory-invariant outputs against an EMA target $\theta^-$,
\begin{equation}
\mathcal{L}_{\mathrm{cm}}
= \mathbb{E}\,\big[\, d\big(
G_\theta(z_i^{\tau},\tau\mid c_i),\;
G_{\theta^-}(z_i^{\tau'},\tau'\mid c_i)\big)\big],
\qquad \tau'<\tau,
\label{eq:cm}
\end{equation}
with $d$ a Huber distance; the combined objective is $\mathcal{L}_{\mathrm{DMD}}+0.5\,\mathcal{L}_{\mathrm{cm}}$. The student itself is a LoRA on the frozen backbone, and the temporal and spatial memory are kept frozen in this stage.

\noindent
The distilled model generates at $4$ sampling steps per chunk with the same $24~\mathrm{fps}$ output, full camera control, temporal memory and spatial memory as its teacher, at a small fraction of the inference cost.

\section{Results}

\subsection{Experimental Setup}

\textbf{Benchmarks.}
We evaluate AlayaWorld on iWorld-Bench~\cite{fang2026iworld}, following its evaluation protocol for the Action Control and Memory Ability tasks.
The evaluation covers three dimensions: Generation Quality, Trajectory Following, and Memory Ability.
These dimensions jointly assess visual quality and consistency, the smoothness and accuracy of action-conditioned trajectories, and the model's ability to preserve spatial and visual consistency along loop-closure trajectories.
In addition, we evaluate AlayaWorld on the standardized WorldMark test suite~\cite{xu2026worldmark} through the World Model Arena, where model outputs generated from identical reference images and action sequences are compared through side-by-side blind human-preference evaluations across Visual Quality, Control Alignment, and World Consistency, with the resulting votes aggregated into Elo ratings.
The evaluation results are publicly available at \url{https://warena.ai/}.

\textbf{Models.}
We compare against representative open-source video world models, including
Cosmos~\cite{agarwal2025cosmos},
HunyuanVideo-1.5 \cite{hunyuanvideo2025},
Yume~1.5~\cite{Mao_2026_CVPR},
Matrix-Game~2.0~\cite{he2025matrix}, and HY-World~1.5~\cite{hyworld2025}.
AlayaWorld is fine-tuned from LTX-2.3~\cite{hacohen2024ltx}, performing autoregressive generation at 720p/540p, where each chunk is produced with four denoising steps and corresponds to roughly one second of video.

\subsection{Quantitative Results}




\begin{table*}[t]
\centering
\caption{
Results on iWorld-Bench across three evaluation dimensions.
All metrics are in $[0,1]$, and higher is better.
}
\label{tab:iworldbench}
\setlength{\tabcolsep}{3.5pt}
\renewcommand{\arraystretch}{1.15}
\resizebox{\textwidth}{!}{%
\begin{tabular}{lccccccc}
\toprule
\textbf{Metric}
& NVIDIA Cosmos
& HunyuanVideo-1.5
& WAN 2.2
& YUME 1.5
& Matrix-Game 2.0
& HY-World 1.5
& \textbf{AlayaWorld} \\
\midrule

\multicolumn{8}{l}{\textbf{Generation Quality}} \\[-1pt]
\hspace{1em}Image Quality
& \underline{0.6778}
& \textbf{0.7128}
& 0.5545
& 0.6232
& 0.4851
& 0.6675
& 0.6620 \\

\hspace{1em}Brightness Consistency
& 0.6952
& 0.7027
& 0.3886
& 0.3810
& 0.2963
& \underline{0.8051}
& \textbf{0.9492} \\

\hspace{1em}Color Temp. Constraint
& 0.7170
& 0.7477
& 0.3411
& 0.4165
& 0.2937
& \underline{0.7819}
& \textbf{0.9379} \\

\hspace{1em}Sharpness Retention
& 0.4363
& 0.5545
& 0.3428
& 0.4023
& 0.4149
& \underline{0.6634}
& \textbf{0.8361} \\

\addlinespace[3pt]
\multicolumn{8}{l}{\textbf{Trajectory Following}} \\[-1pt]
\hspace{1em}Motion Smoothness
& 0.9907
& 0.9908
& 0.9557
& 0.9765
& 0.9848
& \underline{0.9921}
& \textbf{0.9924} \\

\hspace{1em}Trajectory Accuracy
& 0.4955
& 0.6844
& 0.6514
& 0.7113
& 0.7008
& \underline{0.7472}
& \textbf{0.7985} \\

\addlinespace[3pt]
\multicolumn{8}{l}{\textbf{Memory Ability}} \\[-1pt]
\hspace{1em}Memory Symmetry
& 0.3738
& 0.6336
& 0.4480
& 0.5276
& 0.3311
& \underline{0.8481}
& \textbf{0.8871} \\

\hspace{1em}Trajectory Alignment
& 0.6419
& 0.6449
& 0.5703
& 0.5988
& 0.6362
& \underline{0.6776}
& \textbf{0.7018} \\

\bottomrule
\end{tabular}%
}
\end{table*}

Table~\ref{tab:iworldbench} reports the quantitative results on iWorld-Bench across Generation Quality, Trajectory Following, and Memory Ability.
All results are obtained using the distilled autoregressive AlayaWorld model at 480p resolution, matching the resolution of the initial frames provided by the benchmark, with each chunk generated in four sampling steps.
Before inference, we apply an automated, semantics-preserving prompt adaptation procedure to reformulate the benchmark instructions into the prompt style used during training.
Despite the substantially reduced sampling budget, AlayaWorld achieves the best performance on most metrics, demonstrating strong visual consistency, trajectory controllability, and long-term memory.

For Generation Quality, AlayaWorld substantially outperforms the competing methods in brightness consistency, color-temperature constraint, and sharpness retention. These improvements indicate that AlayaWorld effectively mitigates visual drift, illumination fluctuations, and sharpness degradation during autoregressive generation. Although it does not achieve the highest image-quality score, its overall generation quality remains competitive with existing video and interactive world models.

For Trajectory Following, AlayaWorld achieves the best results in both motion smoothness and trajectory accuracy, showing that the generated observations respond smoothly and accurately to the input action trajectories. AlayaWorld also consistently leads on both Memory Ability metrics, demonstrating its ability to preserve visual appearance and spatial structure when revisiting previously observed regions. These results validate the effectiveness of the proposed spatial and temporal memory mechanisms for stable long-horizon interactive generation.


\subsection{Qualitative Results}

We further present qualitative results to demonstrate the interactive generation capabilities of AlayaWorld.
All examples are generated autoregressively by AlayaWorld and cover four representative aspects: camera-controllable navigation, prompt-driven actions, consistent world generation, and long-horizon stability.

\begin{figure}[htbp]
    \centering
    \includegraphics[width=\linewidth]{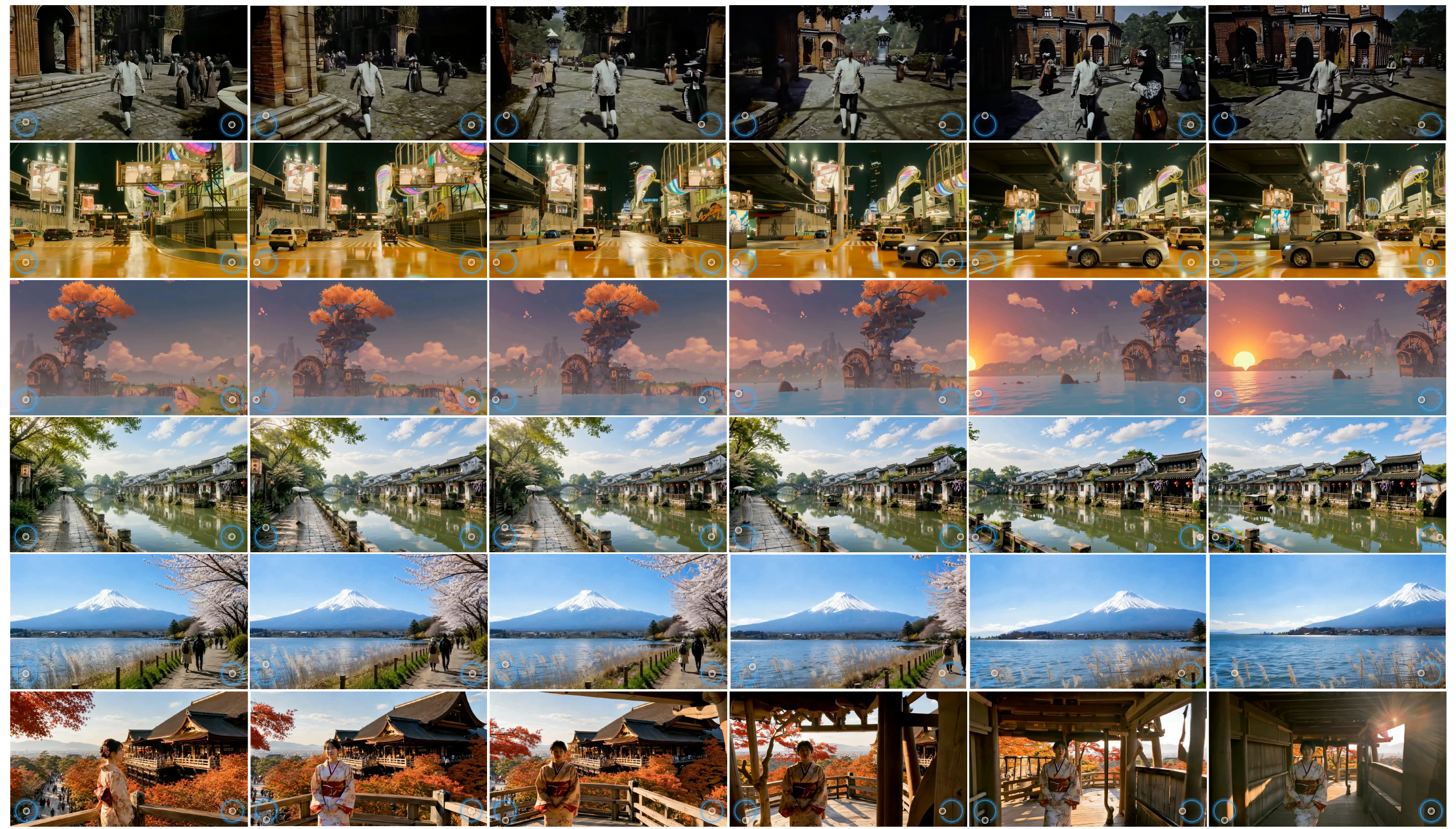}
    \caption{
    Qualitative results of camera-controlled generation.
    AlayaWorld follows diverse camera trajectories while maintaining coherent scene geometry and visual appearance.
    }
    \label{fig:camera}
\end{figure}

Figure~\ref{fig:camera} shows the camera-control results.
Given user-specified camera trajectories, AlayaWorld generates smooth viewpoint changes under diverse translation and rotation commands.
The generated videos follow the intended camera motion while preserving coherent scene geometry, object layout, and visual appearance, demonstrating accurate and stable camera-conditioned navigation.

\begin{figure}[htbp]
    \centering
    \includegraphics[width=\linewidth]{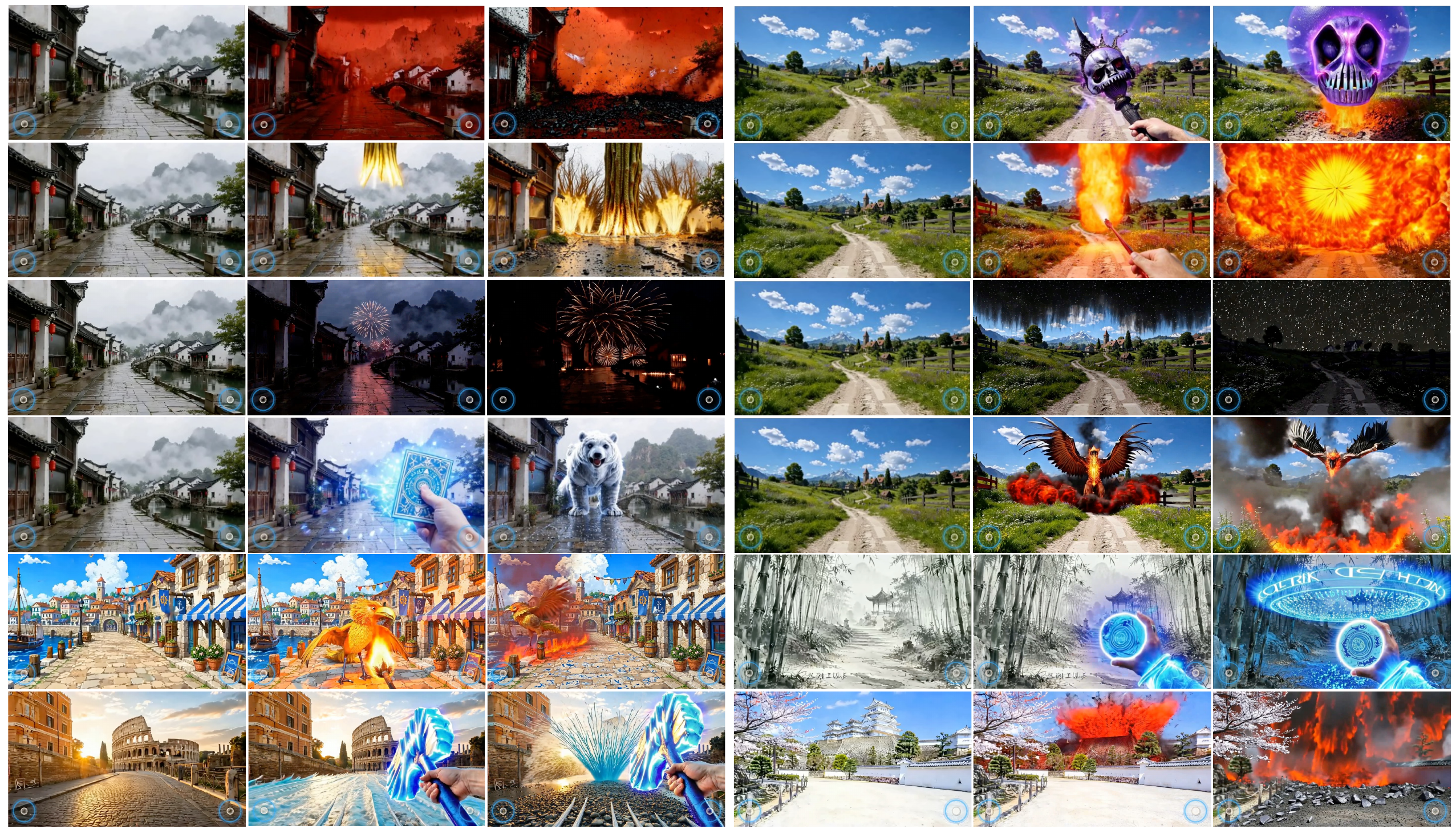}
    \caption{
    Qualitative results of prompt-driven actions.
    AlayaWorld introduces newly requested actions and events during autoregressive generation while preserving the existing scene context.
    }
    \label{fig:magic}
\end{figure}

Beyond navigation, AlayaWorld supports open-ended semantic interactions through dynamically switchable text prompts.
As shown in Figure~\ref{fig:magic}, users can introduce new actions and events, such as spell casting, combat, object appearance, and scene transformation, during an ongoing autoregressive rollout.
The newly specified content emerges naturally in subsequent chunks while the existing scene context and previously generated content remain visually coherent.

\begin{figure}[htbp]
    \centering
    \includegraphics[width=\linewidth]{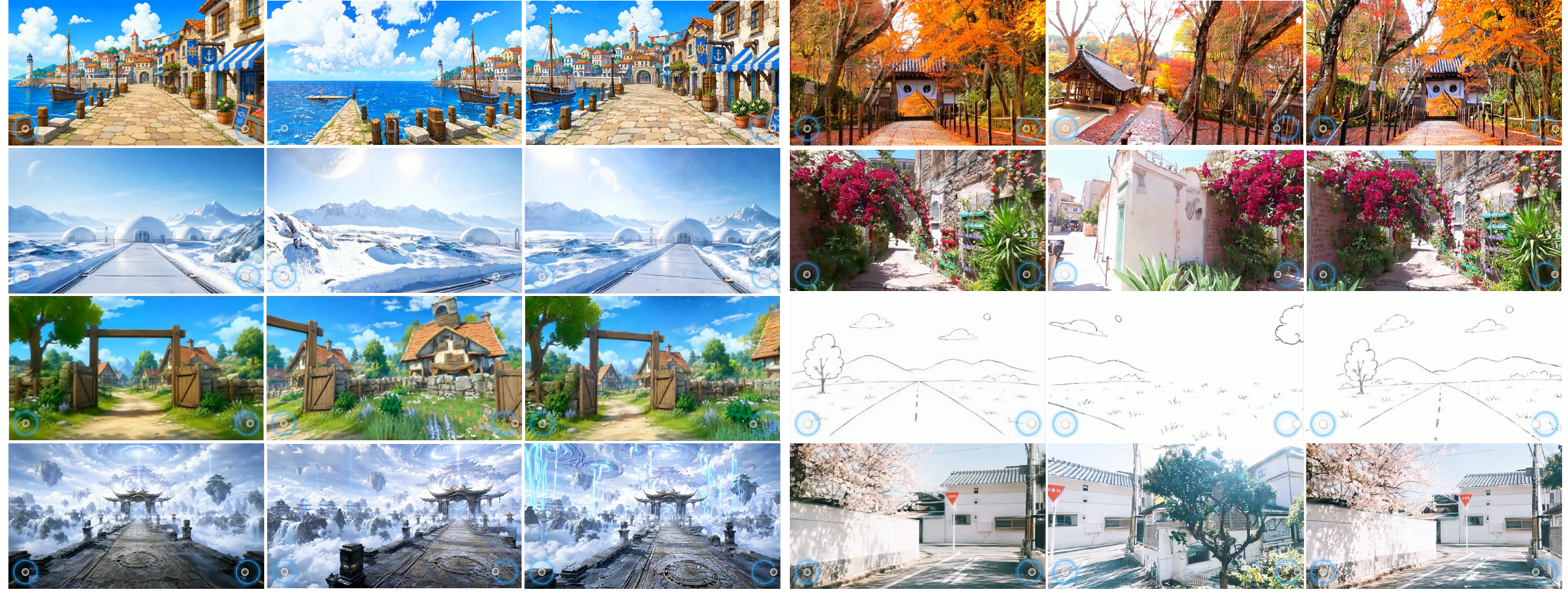}
    \caption{
    Qualitative results of consistent world generation under leave-and-return trajectories.
    AlayaWorld preserves scene structure and visual identity when revisiting previously observed regions.
    }
    \label{fig:consistency}
\end{figure}

Figure~\ref{fig:consistency} demonstrates the world-consistency capability of AlayaWorld under leave-and-return trajectories.
After exploring previously unseen regions, the camera returns to an earlier viewpoint.
AlayaWorld preserves the overall scene layout, object identity, textures, and structural details of the revisited region.
These results indicate that the combination of temporal and geometry-aligned spatial memory effectively supports persistent scene representation beyond the recent context window.

\begin{figure}[htbp]
    \centering
    \includegraphics[width=\linewidth]{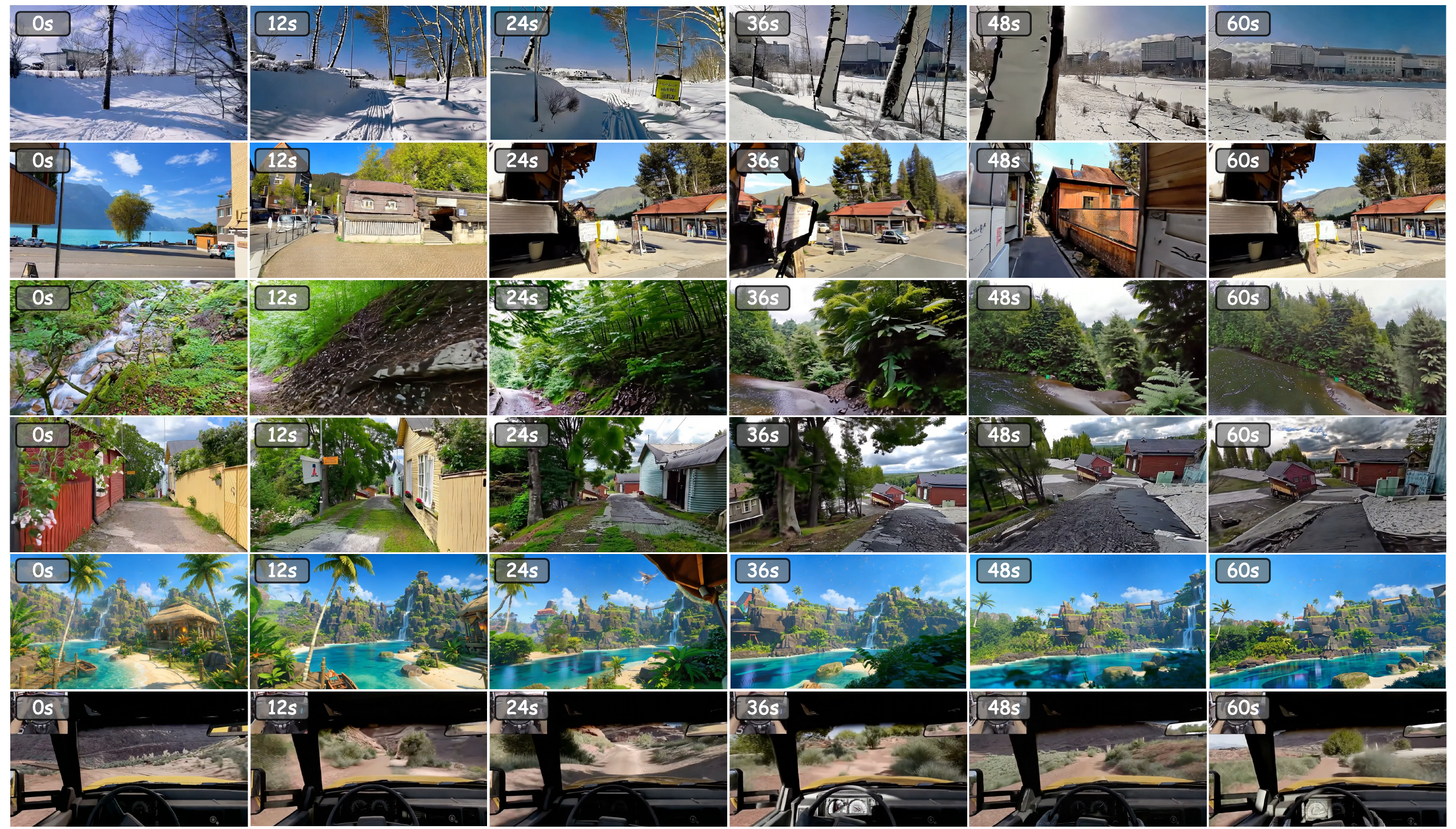}
    \caption{
    Qualitative results of long-horizon autoregressive generation.
    AlayaWorld maintains stable visual quality and coherent scene evolution over extended rollouts.
    }
    \label{fig:long}
\end{figure}

Finally, Figure~\ref{fig:long} presents long-horizon autoregressive generation results.
AlayaWorld maintains coherent scene evolution, stable visual quality, and smooth temporal transitions over extended rollouts.
In particular, the generated videos exhibit limited accumulation of blur, illumination shifts, color drift, and structural degradation, showing the effectiveness of the drift-aware training strategy for stable long-horizon generation.

\section{Conclusion}
\label{sec:conclusion}

We presented \textbf{AlayaWorld}, an open-source interactive long-horizon video world model.
Rather than constructing virtual worlds through conventional game-development pipelines, AlayaWorld explores an alternative paradigm in which interactive environments are synthesized directly from user inputs and evolve continuously through autoregressive generation.

We argue that interactive world modeling is fundamentally defined by four tightly coupled properties: interaction, consistency, stability, and efficiency.
Rather than addressing these properties independently, AlayaWorld integrates them within a unified autoregressive framework in which several design choices simultaneously benefit multiple objectives.
The bounded visual context supports both controllable navigation and persistent world memory, drift-aware training improves long-horizon robustness, and discrete few-step distillation together with short chunk-wise generation enables low-latency interaction while preserving responsiveness to evolving camera trajectories and text prompts.
Experiments on iWorld-Bench demonstrate strong performance across generation quality, trajectory following, and memory ability.
Conceived as a full-stack, open-source, and long-term project, AlayaWorld is intended to provide an extensible foundation for future research on interactive video world models and generative reality.
\section{Contributions and Acknowledgments}

Within each role category, authors are listed in alphabetical order by their first names.

\textbf{Core Lead:}
Kaipeng Zhang

\textbf{Lead:}
Chuanhao Li

\textbf{Core Contributor:}
Chuanhao Li, Kaipeng Zhang, Yifan Zhan, Yongtao Ge, Yuanyang Yin

\textbf{Contributor:}
Jiaming Tan, Kang He, Liaoyuan Fan, Mingliang Zhai, Ruicong Liu, Xiaojie Xu, Xuangeng Chu, Zhen Li, Zhengyuan Lin, Zhixiang Wang, Zian Meng, Zihui Gao

\bibliographystyle{abbrv}
\bibliography{references}

\end{document}